\pdfoutput=1

\documentclass[11pt]{article}

\usepackage[preprint]{acl}
\usepackage{amsmath} 
\usepackage{times}
\usepackage{latexsym}
\usepackage{soul}

\usepackage[T1]{fontenc}

\usepackage[utf8]{inputenc}

\usepackage{microtype}
\usepackage{algorithmic}

\usepackage{inconsolata}

\usepackage{graphicx}

\usepackage{multirow}
\usepackage{verbatim}
\usepackage{longtable}
\usepackage{lscape}
\usepackage{hyperref}
\usepackage{graphicx}
\usepackage{url}
\usepackage{enumitem}
\usepackage{subcaption}
\usepackage{booktabs}
\usepackage{longtable}
\usepackage{array}
\usepackage{tabularx}  
\usepackage{lipsum} 
\usepackage{amssymb} 
\usepackage{pifont}  
\usepackage{makecell}
\usepackage{bbm}
\newcommand{\etal}{\emph{et al.}}

\usepackage[most]{tcolorbox}
\usepackage{xcolor}

\definecolor{TakeBlue}{RGB}{120,170,225}

\tcbset{
  takeaway/.style={
    enhanced,
    breakable,
    colback=TakeBlue!10,
    colframe=TakeBlue,
    coltitle=white,
    fonttitle=\bfseries,
    title=Takeaway,
    boxed title style={colback=blue!60!black},
    left=0.6em,right=0.6em,top=0.4em,bottom=0.4em,
    arc=2mm,
  }
}

\newtcolorbox{takeawaybox}[2][]{%
  takeaway,
  title={#2},
  #1
}
\title{Are LLMs Smarter Than Chimpanzees? \\An Evaluation on Perspective Taking and Knowledge State Estimation}

\makeatletter
\def\thanks#1{\protected@xdef\@thanks{\@thanks
        \protect\footnotetext{#1}}}
\makeatother
\author{Dingyi Yang$^1$, Junqi Zhao$^1$, Xue Li$^2$, Ce Li$^3$, \bf Boyang Li$^1$*\thanks{ *Corresponding Author.}\\
         $^1$College of Computing and Data Science, Nanyang Technological University \\ 
         $^2$University of Science and Technology of China \\ 
         $^3$China University of Mining and Technology\\ 
         \texttt{\{dingyi.yang, junqi.zhao, boyang.li\}@ntu.edu.sg} \\
         \texttt{lixue061219@mail.ustc.edu.cn}, \texttt{celi@cumtb.edu.cn} \\
         }

\begin{document}
\maketitle                
\begin{abstract}
Cognitive anthropology suggests that the distinction of human intelligence lies in the ability to infer other individuals' knowledge states and understand their intentions \cite{tomasello2005understanding}. In comparison, our closest animal relative, chimpanzees, lack the capacity to do so \cite{call2008does,hare2011hominoid}. With this paper, we aim to evaluate LLM performance in estimating other individuals’ knowledge states and their potential actions. We design two tasks to test (1) if LLMs can predict story characters' next actions based on their own knowledge vs. improperly using information unavailable from their perspective, and (2) if LLMs can detect when story characters, through their actions, demonstrate knowledge they should not possess.  Results reveal that most current state-of-the-art LLMs achieve near-random performance on both tasks, and are substantially inferior to humans. We argue future LLM research should place more weight on the abilities of knowledge estimation and intention understanding. The project is available at: \url{https://github.com/DingyiYang/Act-on-Known/}.





\end{abstract}

\section{Introduction}

%


As a flurry of new Large Language Models (LLMs) continues to attain ever-higher scores on performance leaderboards  \cite{deepmind2024imo,xu2025llm}, it is often difficult to shake the feeling that the LLMs are not yet matching human intelligence. On the other hand, it is also increasingly difficult to articulate where exactly the models fall short. What, if any, is the remaining difference between human and machine intelligence?


\begin{figure}[t]
\centering
    \includegraphics[width=.47\textwidth]{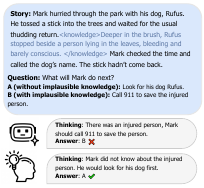}
    \caption{The Knowledge-sensitive Next-action Prediction (KNP) task, where the implausible knowledge is highlighted for easy reading (but not highlighted for the LLMs and the human test-takers). This tests whether LLMs can take the perspective of the story characters and choose actions based on knowledge they can access.}
\label{fig:knowledge_intro}
\end{figure}

To understand the shortcomings of machines, we turn to cognitive anthropological research that attempts to distinguish \emph{homo sapiens} from apes and other animals. A popular view there is that humans possess an advanced theory of mind (ToM) \cite{premack1978does} but other animals do not \cite{Poninelli-2003:Chimpanzee-minds,Suddendorf-2008:foresight,Krupenye-2019:ToM-Review}. That is, humans are uniquely capable of understanding and emulating the mental states of other individuals.

\begin{table*}[t]
\centering
\fontsize{5.7}{7.2}\selectfont
\renewcommand{\arraystretch}{0.88}
\setlength{\tabcolsep}{3.8pt}
\begin{tabular}{@{}lccccccccc@{}}
\toprule
& \multicolumn{3}{c}{\textbf{Ability Tested}}
& \multicolumn{2}{c}{\textbf{Knowledge Domain}}
& \multirow{3}{*}{\textbf{\#Samples}}
& \multirow{3}{*}{\textbf{Story Length}}
& \multicolumn{2}{c}{\textbf{Performance}} \\
\cmidrule(lr){2-4}\cmidrule(lr){5-6}\cmidrule(lr){9-10}
& \makecell[c]{Knowledge State\\[-2pt] Tracking}
& \makecell[c]{Potential Action\\[-2pt]Estimation}
& \makecell[c]{In-Context\\[-2pt]Localization}
& \makecell[c]{Knowledge\\[-2pt]Domain}
& \makecell[c]{Knowledge\\[-2pt]Creation}
& & & Random & GPT-5.4 (high) \\
\midrule
\multicolumn{10}{l}{\bfseries\emph{Theory-of-Mind Benchmarks}} \\
ToMi \cite{le2019tomi} & \checkmark & - & - & Location & Rule-Based & 6K & 40 words & 0.50 & 0.90 \\
FANTOM \cite{kim2023fantom} & \checkmark & - & - & 3 Types & LLM & 10K & 537 words & 0.50 & 0.76 \\
T4D \cite{zhou2023t4d} & \checkmark & \checkmark & - & Location & LLM & - & - & - & - \\
SIMPLETOM (BehaviorQA) \cite{gu2026simpletom} & \checkmark & \checkmark & - & 10 Types & LLM & 1.1K & 38 words & 0.50 & 0.77 \\
CharTOM-QA (BeliefQA) \cite{zhou2025essence} & \checkmark & \checkmark & - & General & Classic Novel & 201 & 2K words & 0.25 & 0.65 \\
\midrule
\multicolumn{10}{l}{\bfseries\emph{Benchmarks Proposed in This Paper}} \\
KNP (Ours) & \checkmark & \checkmark & - & General & Human \& LLM & 500 & 93 words & 0.50 & 0.61 \\
IKD (Ours) & \checkmark & \checkmark & \checkmark & General & Human \& LLM & 1K & 601 words & 0.25 & 0.53 \\
\bottomrule
\end{tabular}
\caption{Comparison between existing datasets. KNP requires models to \textit{track characters’ knowledge states} and \textit{estimate their potential next actions}. IKD requires models to do so for each event in context, and to \textit{localize any implausible action}. Compared to datasets that rely on limited LLM-generated scenarios or classic novels prone to data leakage, our datasets cover broader situations and require richer commonsense understanding of the real world.}
\label{table:dataset-comparison}
\end{table*}

Among these, arguably the most influential theory comes from Tomasello \etal, who posit that the \emph{one} fundamental distinction between humans and other animals is the ability to \textit{understand and share the intentions of others} (USI) \cite{tomasello1999cultural}. \citet{tomasello2005understanding} propose that USI contains a few components: First, one must recognize that other individuals are autonomous and have independent mental states; otherwise it may be assumed that everyone knows and wants exactly the same things. Second, one needs to understand that other individuals choose their actions based on the knowledge and actions available to them. Interestingly, experiments suggest that apes are unable to track what other apes know and believe \cite{call2008does,hare2011hominoid}. It stands to reason that chimpanzees cannot infer intentions of others from reliable estimations of their knowledge states. 



%


Therefore, a natural question to ask is: can LLMs infer the knowledge states of human actors? Using textual stories, we propose two probing tasks: (1) \textbf{Knowledge-sensitive Next-action Prediction} (KNP) (Fig. \ref{fig:knowledge_intro}), which tests whether LLMs can predict actions that story characters will likely take, depending on their own knowledge. (2) \textbf{Implausible Knowledge Detection} (IKD) (Fig. \ref{fig:data_construction}), which tests whether LLMs can detect when a story character acts on knowledge they should not possess.



As shown in Figure \ref{fig:data_construction}, we construct two benchmark datasets, one for each task. We collect and summarize human-written stories from the internet\footnote{\url{www.reedsy.com} and \url{www.short-story.me}}. For KNP, we identify or create \footnote{If no such knowledge exists in the original story, we use LLM to introduce one.} a knowledge statement that cannot be plausibly known by a story character, but has important implications for their next action. The LLM being tested must predict the next action of the character, and the correct choice is that the character will act without the knowledge. The dataset consists of 500 QA pairs in 5 genres.

For the IKD task, we rewrite the story to introduce a plot hole \cite{ahuja2025finding}. In the plot hole, a story character uses knowledge they cannot possibly possess in their action. The LLM is required to detect if a plot hole exists in a given story. The benchmark consists of 500 correct stories and 500 erroneous stories. Both KNP and IKD stories were validated by human annotators. Section \ref{sec: dataset} contains more details.

Our proposed benchmarks demand a more advanced capacity for tracking knowledge states and understanding how they affect actions in the real world (Table \ref{table:dataset-comparison} shows comparisons with existing datasets). They are highly adversarial to state-of-the-art LLMs. On the KNP task, humans with zero training and zero demonstrations achieve an accuracy of 92\%, 23 percentage points (pp) above the best LLM, Claude-4.5-Opus. GPT-5.4 achieves only 61\%, whereas GPT-4o performs at random chance. On the IKD task, Claude-4.5-Opus achieves 68\% classification accuracy and 0.59 IoU for localization, whereas other models remain at or below roughly 60\% accuracy.  The zero-shot human baseline reaches 76\% accuracy. The surprising failures of LLMs in understanding seemingly simple stories highlight their deficiency in the critical ability of estimating knowledge states and potential actions.


This paper makes the following contributions: 
\begin{itemize} 
\setlength\itemsep{0pt}
    \item Inspired by cognitive research, we propose an important dimension of LLM evaluation: the ability to understand and share intentions (USI) of humans. As USI can be difficult to measure directly, we propose to evaluate a prerequisite skill: knowledge state tracking and potential action estimation. 
    \item We build simple and short tests that push most state-of-the-art LLMs (except the latest Claude and Gemini models) to marginally above random chance. The finding is interesting by itself and highlights the need for further research on USI. 
\end{itemize}

\section{Related Works}\label{sec:related_work} 
\paragraph{Theory of Mind (ToM) in LLMs.} 
Fundamental to social intelligence, Theory of Mind refers to the ability to recognize the fact that other people possess distinct mental states and to estimate such states \cite{nguyen2025survey}. Most studies that test ToM in LLMs focus on the dimension of beliefs and knowledge \cite{kosinski2024evaluating,kim2023fantom,wu2023hi,nematzadeh2018evaluating, xu2024opentom}, others cover intentions \cite{staab2023beyond,zhou2023cast} and emotions \cite{wu2024coke}.

In this work, we focus on the dimension of knowledge. Most existing works face two limitations: \textbf{1)} they explicitly ask about knowledge state understanding, like in ToMi \cite{le2019tomi}, Hi-ToM \cite{wu2023hi}, and FANTOM \cite{kim2023fantom}. For example, given a story ``Sally is in one room; in another room, Anne puts milk in a box,'' they ask ``What does Sally think is in the box?'', which directly cues LLMs to perform knowledge tracking. In contrast, testing the LLMs on the \textit{action implications of knowledge} (our setting) is harder, using questions like ``Sally wants to drink milk. What would she do? A: Go to the box in another room. B: Try to find milk.'' The inference that the knowledge state is critical in this decision appears to be rather difficult for LLMs (see the first ablation in \S \ref{sec:analysis}) but is vital for USI. \textbf{2)} Many works employ false-belief contexts, where the story presents a world state change (e.g., ``Sally puts milk in a basket and leaves; Anne moves it to a box''), like in T4D \cite{zhou2023t4d} and Open-TOM \cite{xu2024opentom}. This setup provides two alternative world states, implicitly letting LLMs choose between them. In contrast, our stories present only an unchanging fact that the character cannot know. If LLMs fail to understand that one's action is based on their own knowledge, they will directly use this inaccessible fact. This design more rigorously tests whether LLMs understand how perspective-specific knowledge affects action.

Additionally, compared to datasets built from standardized templates \cite{le2019tomi} or limited scenarios \cite{gandhi2023understanding,chen2024tombench,gu2026simpletom}, our dataset is drawn from diverse human-written stories, which cover a broader range of realistic scenarios and offer a more out-of-distribution assessment for ToM. Although CharToM-QA \cite{zhou2025essence} also targets general scenarios, it relies on classic novels and is subject to serious data leakage concerns. Moreover, the lengthy context (over 2K words) makes it unclear whether the challenge stems from context length or ToM understanding. Our KNP task, by contrast, features much shorter natural narratives (93 words on average), yet LLMs still show near-random performance.

\paragraph{Narrative Understanding.} 
Many works focus on basic story element understanding, such as NarrativeQA \cite{karpinska2024one}, BookQA \cite{angelidis2019bookqa}, and FairyTaleQA \cite{xu2022fantastic}. FlawedFictions \cite{ahuja2025finding} investigates if LLMs can detect plot holes that may require complex logical reasoning. Our IKD task uses a similar setting but focuses on plot holes related to characters' knowledge states.

\section{Tasks of Implausible Knowledge}
We define implausible knowledge as knowledge that an individual's physical location, social status, or personal experience does not plausibly grant them. For instance, if a story character knows a stranger's name at first glance, we would consider it implausible\footnote{Unless, of course,  the story already established that the character has magic or some mind-reading device.}. Other examples include knowledge of private thoughts of others, or knowledge about events that happened in a far away location, in the character's absence, or in the future.

We design two tasks to evaluate LLMs' understanding of knowledge. The KNP task (Sec \ref{sec:task1}) focuses on the prediction of next actions \emph{ex ante}, whereas the IKD task (Sec \ref{sec:task2}) focuses on the detection of actions that depend on implausible knowledge \emph{ex post}. As such, these two tasks test two complementary aspects of the understanding of implausible knowledge. 


\subsection{Task 1: Knowledge-sensitive Next-action Prediction}\label{sec:task1}
In this task, each story contains an implausible piece of knowledge $\mathcal{K}$ that the story character $C$ should not know, but $\mathcal{K}$ has strong implications on the next action of $C$. For example, $C$ is looking for ancient treasures and $\mathcal{K}$ is the location of the treasure. Upon learning $\mathcal{K}$, $C$ would go to the location immediately. Without $\mathcal{K}$, it would be implausible for $C$ to go to the exact right location without proper justification. 

After reading the story, the LLM is asked to choose the more plausible next action for story character $C$. There are two choices, an action that $C$ would normally take without knowing $\mathcal{K}$, and an action $C$ would take knowing $\mathcal{K}$. Clearly, the first action is the correct choice. We report the accuracy.

\subsection{Task 2: Implausible Knowledge Detection} \label{sec:task2}
This task evaluates if an LLM can determine whether a story character possesses knowledge inaccessible to them given the narrative context. Formally, let $S$ be a narrative consisting of a sequence of events $E = \{e_1, e_2, ..., e_n\}$. An erroneous story with implausible knowledge $\mathcal{K}$ satisfies two conditions: (1) there is an event $e_k$ where a story character $C$ either states the information $\mathcal{K}$ or takes an action premised on $\mathcal{K}$, and (2) it is highly unlikely for $C$ to know $\mathcal{K}$ at the time of $e_k$.

The task contains two subtasks. In the binary classification subtask, the LLM classifies each story into logically consistent (1) or inconsistent (0). We report the classification accuracy. Given a story $x_i$, a predicted class $\hat{y}_i$, and the ground-truth label $y_i$, the accuracy is simply $\sum_i^N \mathbbm{1}(\hat{y}_i=y_i) / N$, where $N$ is the number of stories and $\mathbbm{1}(\cdot)$ is the indicator function. 

In the localization subtask, if the LLM determines the story to be logically inconsistent, it must localize the error by identifying the error-inducing sentences $s_{err}$. We prompt the LLM to predict the error sentences $\hat{s}_{err}$ in words, and compute the intersection-over-union (IoU) with the ground truth at the word level.  The localization metric $\mathrm{l}_i$ for the $i$-th story is calculated as: 
\[
    \mathrm{l}_i=
\begin{cases}
\mathbbm{1}(\hat{y}_i=1), \;\text{if}\; y_i=1 \; \text{(no error)}\\
\mathbbm{1}(\hat{y}_i=0) \, \mathrm{IoU}(s_{err},\hat{s}_{err}), \;\text{if}\; y_i=0 \; \text{(error)}\\
\end{cases}
\]
The dataset-level localization metric is computed as the average of $\mathrm{l}_i$. 




\begin{figure*}[t]
\centering
    \includegraphics[width=0.95\textwidth]{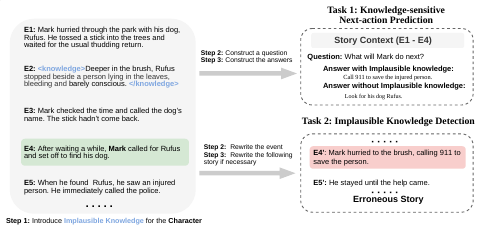}
    \caption{Our dataset construction process (Section \ref{sec: dataset}). Using the original story, we first identify \textit{implausible knowledge} that a \textit{character} cannot possess at a given \textit{event}, and that would affect the character's action. If no such knowledge exists, we prompt the model to introduce one. Then: 1) For the KNP task, we construct a question about a knowledge-sensitive action, creating two answer options—one without the implausible knowledge and one with the knowledge; 2) For the IKD task, we rewrite the original event to generate an erroneous story.}
\label{fig:data_construction}
\end{figure*}

\section{Dataset Construction} \label{sec: dataset}
\subsection{Story Collection}
We collect 500 open-source stories from \url{www.reedsy.com} and \url{www.short-story.me}, covering five popular genres of romance, fantasy, kids, mystery, and science fiction. To avoid copyright issues, we perform automatic summarization of all collected stories (see Fig. \ref{table:story_summary} for the summarization prompt), yielding the raw original stories $S_{ori}$.

\subsection{Data Construction}
As depicted in Figure \ref{fig:data_construction}, we propose a data construction process that combines LLM generation and human validation and refinement. For LLM generation, we use GPT-5 with high reasoning for all steps. Detailed prompts are in the Appendix.
\paragraph{Introducing Implausible Knowledge.} 
Given an original story, we first identify a piece of referenced knowledge that satisfies two criteria: (1) at the time of an event $e_k$, a character cannot plausibly possess this knowledge; and (2) the knowledge nevertheless could influence the character’s actions in $e_k$. If no suitable instance of such knowledge is present, we prompt the model to generate one and insert it into the narrative in a natural manner.

\paragraph{Next-action Question Construction.}
Given the implausible knowledge $\mathcal{K}$, character C, and event $e_k$, we use the story context preceding $e_k$ as the input to the KNP task. We then construct a question about the character’s next action, where the answer depends on whether $C$ knows $\mathcal{K}$. We provide two candidate answers: a correct option and a distractor. The question-generating LLM is prompted to ensure that, assuming that $C$ does not know $\mathcal{K}$, the action without $\mathcal{K}$ is clearly the correct choice and the distractor is not independently plausible for other reasons. 

Human annotators check for the correctness of the answers. For example, suppose $\mathcal{K}$ is “the girl whom $C$ loves thinks he is weird,” and $C$ is characterized as very shy. If the correct option is ``he finds the courage to confess'' and the distractor is ``he does not have the nerve to confess'', the distractor may remain plausible solely due to shyness rather than being caused by $\mathcal{K}$. In this case, human annotators may choose to rewrite the options themselves or regenerate using the LLM. The final refined dataset contains 500 QA pairs for the KNP task, with stories averaging 93 words in length.

\paragraph{Erroneous Story Construction.} The original stories $S_{ori}$ do not contain plot holes. To obtain the erroneous stories for the IKD task, we rewrite event $e_k$ under the counterfactual assumption that the character possesses the identified implausible knowledge, yielding a modified event $e_k'$. The LLM is prompted to produce a rewrite that is fluent and coherent while introducing as few changes to the original event as possible. At the same time, the inconsistency introduced by $e_k'$ should not be explainable within the story world. In addition, if rewriting $e_k$ affects subsequent events, the LLM is asked to make the minimal adjustments necessary to preserve narrative continuity after $e_k'$.

During human validation, the annotators re-check the validity of $e_k'$ and overall continuity. For example, cases in which $e_k'$ can be justified by ``the character may obtain knowledge with magic power'' or ``the character is lying about their possession of the implausible knowledge'' are not considered as errors. Human annotators regenerate $e_k'$ until the requirements are met. 

Further, the human annotators are asked to remove explicit explanations that character $C$ does not know $\mathcal{K}$ in the story. The LLM being tested should reason about character knowledge states using commonsense, rather than hints in the text. 

Combining the 500 erroneous and 500 original correct stories\footnote{Original stories are also verified and refined by human annotators to remove any possible logic errors.}, this yields the dataset for the IKD task, with an average length of 601 words.

Overall, approximately 40\% of the samples in both tasks are refined by annotators. Additional details about the annotators and data refinement requirements are provided in the Appendix.


\begin{table*}[t]
\centering
\fontsize{8.5}{11.2}\selectfont
\begin{tabular}{l*{6}{c}}
\toprule
    \multirow{3}{*}{\textbf{}} & 
    \textbf{KNP} &
    \multicolumn{5}{c}{\textbf{Implausible Knowledge Detection (IKD)}} \\
    
    \cmidrule(lr){2-2} \cmidrule(lr){3-7}
     & 
    \textbf{Action Choice} &
    \multicolumn{3}{c}{\textbf{Binary Classification}} & 
    \multicolumn{2}{c}{\textbf{Localization}} \\
    
    \cmidrule(lr){2-2} \cmidrule(lr){3-5} \cmidrule(lr){6-7}
     & \textbf{Accuracy} & \textbf{Pos Cls} & \textbf{Neg Cls} & \textbf{Full Cls} & \textbf{Neg Loc} & \textbf{Full Loc} \\
    \midrule
\multicolumn{7}{c}{\emph{Open-weight Models}} \\
Qwen3-8B         & 0.50 & 0.58&0.31&0.45&0.04&0.31 \\
Qwen3-8B (think) & 0.49 & 0.69&0.32&0.51&0.05&0.37 \\
Qwen3-32B        & 0.49 & 0.78&0.23&0.51&0.04&0.41 \\
Qwen3-32B (think)& 0.50 & 0.65&0.45&0.55&0.10&0.38 \\
Qwen3-Max (think)& 0.53 & 0.86&0.24&0.55&0.07&0.47 \\
LLaMA3.1-8B      & 0.51 & 0.14&0.64&0.39&0.11&0.13 \\
LLaMA3.1-70B     & 0.55 & 0.41&0.60&0.51&0.08&0.25 \\
LLaMA3.3-70B     & 0.54 & 0.69&0.36&0.53&0.05&0.37 \\
Kimi-K2.5-Instruct & 0.56 & 0.71&0.51&0.61&0.27&0.49 \\
Kimi-K2.5-Thinking & \textbf{0.57} & 0.74&0.49&\textbf{0.62}&0.26&0.50 \\
DeepSeek-V3.1    & 0.47 & 0.40&0.65&0.53&0.16&0.28 \\
DeepSeek-R1-V3.1 & 0.55 & 0.76&0.46&0.61&0.22&0.49 \\
DeepSeek-V3.2    & 0.50 & 0.68&0.50&0.59&0.20&0.44 \\
DeepSeek-R1-V3.2 & \textbf{0.57} & 0.94&0.27&0.61&0.16&\textbf{0.55} \\
\midrule
\multicolumn{7}{c}{\emph{Closed-weight Models}} \\
GPT-4o           & 0.49 & 0.92&0.10&0.51&0.03&0.48 \\
GPT-5-mini       & 0.52 & 0.83&0.18&0.51&0.06&0.45 \\
GPT-5 (low)      & 0.57 & 0.81&0.37&0.59&0.18&0.50 \\
GPT-5 (medium)   & 0.58 & 0.80&0.33&0.57&0.15&0.48 \\
GPT-5 (high)     & 0.58 & 0.85&0.30&0.58&0.13&0.49 \\
GPT-5.4 (high)   & 0.61 & 0.93&0.27&0.60&0.12&0.53 \\
Gemini-3-Pro     & 0.67 & 0.28&0.86&0.57&0.47&0.38 \\
Claude-4.5-Sonnet& 0.66 & 0.35&0.82&0.59&0.29&0.32 \\
Claude-4.5-Opus  & \textbf{0.69} & 0.88&0.48&\textbf{0.68}&0.30&\textbf{0.59} \\
\midrule
Zero-shot Human  & 0.92 & 0.82&0.70&0.76&0.61&0.72 \\
\bottomrule
\end{tabular}
\caption{Main results on KNP and IKD tasks. For IKD, we show the classification and localization performance on the positive set (correct stories), negative set (erroneous stories), and the full set.}
\label{table:main_results}
\end{table*}


\section{Experiments}\label{sec:results}

\paragraph{LLM Tested.}
We test several models across two categories: closed-source models (Gemini-3-Pro, GPT-5.4, GPT-5, GPT-5-mini, GPT-4o, Claude-4.5-Sonnet, and Claude-4.5-Opus) and open-source models (LLaMA3.3-70B \cite{grattafiori2024llama3herdmodels}, LLaMA3.1-70B \cite{dubey2024llama}, LLaMA3.1-8B, Qwen3-MAX, Qwen3-32B \cite{yang2025qwen3technicalreport}, Qwen3-8B, DeepSeek-Chat, DeepSeek-R1 (v3.1 and v3.2) \cite{liu2025DeepSeek}, Kimi-K2.5-Instruct \cite{team2026kimi}, and Kimi-K2.5-Thinking.

\paragraph{Settings.} We test the LLMs using the zero-shot setting. Since the task requires reasoning about knowledge states, we always ask the models to generate their thinking process and output the answer. For temperature and top-p settings, we apply the default settings for all open-source models and closed-source APIs, then average scores across 3 runs. For the KNP task, we randomly shuffle the position of correct and distractor answers to avoid positional bias in LLMs.

\paragraph{Human Performance.} We recruited 2 annotators with sufficient English proficiency and gave them the same instructions we provided to the LLMs. For each task, the annotators solved 50 questions.

\subsection{Main Results} The results are shown in Table \ref{table:main_results}. For knowledge-sensitive action prediction (KNP), most models perform near-randomly — especially open-source models. Among leading models, Claude-4.5-Opus and Gemini-3-Pro achieve the highest performance at around 0.7. Analysis of their reasoning processes reveals that they consider knowledge states when predicting character actions (Case shown in Fig. \ref{fig:case3}). However, their understanding of knowledge states still needs improvement, revealing a gap with human performance of 0.92.

For logic error detection (IKD), many leading models perform only slightly better than random in the \textit{binary classification task}. For instance, GPT-5 (high) achieves 0.58, DeepSeek-R1 reaches 0.61, and Claude-4.5-Opus shows the highest performance at 0.68. In the \textit{localization task}, performance is even lower—DeepSeek-R1 achieves 0.55, and Gemini-3-Pro reaches 0.38. An interesting finding is that different models also exhibit distinct judgment tendencies: GPT-4o shows a strong bias toward determining stories are logically consistent, while Gemini-3-Pro shows the opposite tendency. As a result, although some models display high performance on either the positive set (stories with no errors) or the negative set, the overall performance is poor. Human performance is also not particularly high, largely due to the lengthy context. If we highlight the events containing the error and the relevant knowledge, their localization performance on the negative set increases from 0.61 to 0.86.

\paragraph{Effect of Scale.} Comparing LLMs with different parameter counts, such as LLaMA-3.1 (8B vs. 70B) and Qwen3 (8B, 32B, MAX), we see that model scaling improves performance only slightly. This suggests that knowledge estimation and tracking may not be easily solved by scaling alone.


\paragraph{Effect of Extended Reasoning.} We explore whether different model series show improvement with extended thinking processes. For instance, we compare DeepSeek-chat and DeepSeek-R1, Qwen3's thinking and non-thinking modes\footnote{For non-thinking mode, the model still outputs a short chain of thought.}, as well as GPT-5's different reasoning levels. However, extended reasoning does not lead to clear improvement. This highlights that knowledge estimation and tracking cannot simply be resolved through test-time scaling and requires specialized training. 

\begin{takeawaybox}{Takeaway}
 Most state-of-the-art LLMs show near-random performance in knowledge state tracking and potential action estimation. Test-time scaling does not show a clear advantage.
\end{takeawaybox}

\subsection{Further Analysis and Ablation} \label{sec:analysis}
This section presents an in-depth analysis of LLMs' understanding of knowledge.

\begin{figure*}[t]
    \centering
    \includegraphics[width=\linewidth]{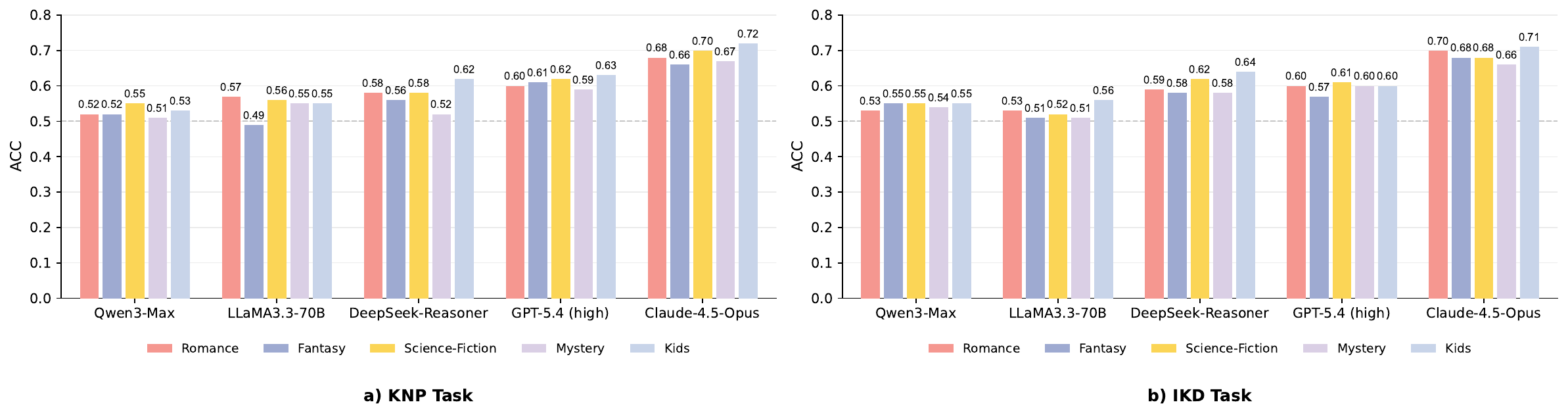}
    \caption{Performance on different genres.}
    \label{fig:genres}
\end{figure*}

\paragraph{Effects of Genre.}
We examine whether narrative style affects performance. We compare five genres in our benchmark dataset: romance, fantasy, science fiction, kids, and mystery. As shown in Figure \ref{fig:genres}, different genres show no clear performance difference across either task. The fantasy and mystery genres show slightly lower performance—possibly because models tend to excuse errors as explainable within the fantastic and some mysterious settings.\footnote{The stories explicitly state that the characters in our questions do not possess magic power (so they cannot use magic to obtain implausible knowledge).}

\paragraph{Ablation: Focusing LLMs on Character Knowledge.}

In the basic setting used in Table \ref{table:main_results}, the LLMs are not told that they need to track knowledge states of story characters. In order to test if the LLMs would perform differently if they know the ability being tested, in this experiment, we explicitly instruct them to focus on the specific knowledge states of story character. 

\begin{figure*}[t]
    \centering
    \includegraphics[width=\linewidth]{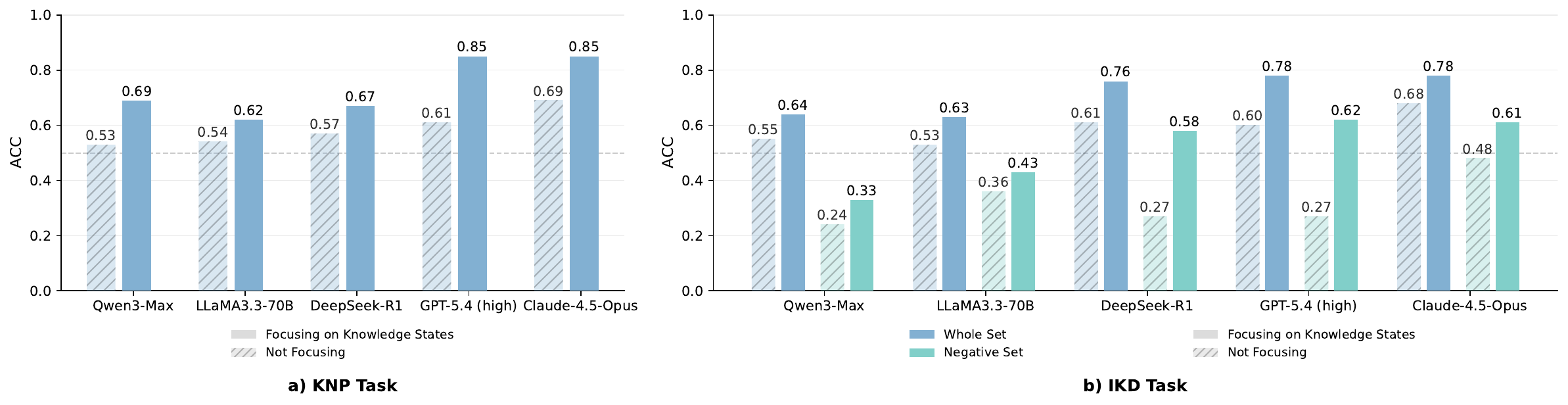}
    \caption{Ablation: Focusing LLMs on Character Knowledge.}
    \label{fig:explicit_knowledge}
\end{figure*}

The results are shown in Figure \ref{fig:explicit_knowledge}. On average, the explicit prompt to pay attention to character knowledge improves accuracy by 12.6 pp in IKD (Whole Set) and 14.8 pp in KNP, with GPT-5.4 achieving the highest boost. This suggests the LLMs do not yet understand the dependency between actions taken by humans and their knowledge, which is a critical component of USI according to \citet{tomasello2005understanding}. After being prompted, the best accuracy is 85\% on KNP, still falling short of the performance of the human participants who did not receive this prompt.

\paragraph{IKD Localization Ablation: Isolating Knowledge Tracking.}
The IKD localization task is the most complex and has the lowest performance. Doing this correctly requires making a sequence of decisions correctly: (1) identifying the statement of the implausible knowledge. (2) identifying the action of the story character premised on the implausible knowledge, (3) knowing that knowledge inconsistency needs to be checked, and (4) performing knowledge tracking correctly to classify the story and localization. This makes it difficult to interpret the results of IKD. 
To isolate the effects of Step \#4, we build the following ablated versions of the IKD localization task. 

\noindent \textbf{Ablation 1:} We explicitly highlight the erroneous event and instruct the model to identify whether it contains a logical inconsistency. Compared to the full task, this setting removes the need to scan and reason over all events, reducing the problem to detecting a logical error within a single, pre-identified event.

\noindent \textbf{Ablation 2:} We further highlight both the erroneous event and the event that introduces the implausible knowledge for the character, and ask the model to detect the logical inconsistency. This setting eliminates the challenge of locating the relevant background knowledge, requiring the model only to reason about the inconsistency between the two given events.

\noindent \textbf{Ablation 3:} Building on Ablation 2, we additionally instruct the model to explicitly focus on implausible character knowledge states. This represents the simplest setting, isolating the core capability under evaluation: whether the model understands that a character cannot possess certain knowledge given the narrative context.

The results are shown in Table \ref{tab:ablation_models}. We find that for all models, the first two ablations do not bring clear improvement. However, in Ablation 3—where models are asked to focus on knowledge states—performance shows a clear increase. This aligns with our previous experimental findings: models do not pay much attention to knowledge states in their default setting. Yet even under this condition, performance remains modest for most models, highlighting the need for future improvements.
\begin{takeawaybox}{Takeaway}
  (1) Even when other required skills, such as retrieval in a long context, are removed from the problem, the LLMs still perform worse than humans. \\
  (2) Performance improves when LLMs are asked to focus on knowledge, suggesting LLMs do not understand the dependency between actions taken by humans and their knowledge, which is critical for USI \cite{tomasello2005understanding}.
\end{takeawaybox}

\begin{table}[t]
\centering
\fontsize{7.6}{9.8}\selectfont

\begin{tabular}{lcccc}
\toprule
     & \multicolumn{4}{c}{\textbf{Localization (Negative Story)}} \\
    \cmidrule(lr){2-5}
    \textbf{Model} & \textbf{Default} & \textbf{Abl. 1} & \textbf{Abl. 2} & \textbf{Abl. 3} \\
\midrule
Qwen3-8B (think) & 0.05&0.12&0.14&0.33  \\
Qwen3-32B (think)  & 0.10&0.34&0.29&0.60  \\
Qwen3-Max (think)  &0.07&0.28&0.33&0.69  \\
LLaMA3.1-8B  & 0.11&0.20&0.23&0.28 \\
LLaMA3.1-70B & 0.08&0.29&0.39&0.67  \\
LLaMA3.3-70B   & 0.05&0.18&0.3&0.55 \\
Kimi-K2.5-Instruct & 0.27&0.47&0.54&0.70  \\
Kimi-K2.5-Thinking & 0.26&0.46&0.54&0.71  \\
DeepSeek-V3.2 & 0.20&0.36&0.41&0.78   \\
DeepSeek-R1-V3.2  & 0.16&0.26&0.30&0.59 \\
GPT-4o   & 0.03&0.12&0.22&0.62 \\
GPT-5.4 (high)& 0.12&0.22&0.30&0.61 \\
Gemini-3-Pro & 0.47&0.68&0.73&0.80 \\
Claude-4.5-Opus   & 0.30&0.45&0.48&0.62  \\
\midrule
Human          & \textbf{0.61} & - & \textbf{0.86} & - \\
\bottomrule
\end{tabular}
\caption{Ablation studies on the IKD-localization task (Neg set), which isolate the ability to estimate knowledge states from long-context understanding.}
\label{tab:ablation_models}
\end{table}

\paragraph{KNP: Binary Choice vs. Open-Ended Generation.}
For the Knowledge-sensitive Next-action Prediction task, the default setting is a binary choice task. We explore whether the provided answers serve as hints or distractors for the models. To investigate this, we conduct an open-ended prediction experiment: without given option answers, we first generate the open-ended answer, then apply the LLM-as-a-Judge method \cite{li2024llms} for open-ended evaluation. We provide GPT-5 with the story, question, answer options, and generated answer. If it aligns with the correct option, it's correct; if aligns with the incorrect option, it's wrong. Answers aligning with neither are filtered out—7\% on average.

The results are shown in Table \ref{tab:open-ended}. We observe that the two question formats do not create substantial differences in the scores, demonstrating that the conclusions based on binary choice setting are robust. 

\begin{table}[t]
\centering
\fontsize{7.6}{9.8}\selectfont
\begin{tabular}{lcc}
\toprule
    \textbf{Model} & \textbf{Binary Choice} & \textbf{Open-Ended} \\
\midrule
Qwen3-8B (think)      & 0.49&0.53\\
Qwen3-32B (think)     & 0.49&0.51\\
Qwen3-Max (think)   & 0.53&0.53\\
LLaMA3.1-8B           & 0.51 & 0.52 \\
LLaMA3.1-70B          &0.55&0.57\\
LLaMA3.3-70B          &0.54&0.58\\
Kimi-K2.5-Instruct      &0.56&0.56 \\
Kimi-K2.5-Thinking      &0.57&0.56 \\
DeepSeek-V3.2         & 0.49&0.51   \\
DeepSeek-R1-V3.2      & 0.57&0.58\\
GPT-4o                & 0.49&0.52\\
GPT-5.4 (high)          &0.61&0.62\\
Gemini-3-Pro          & 0.67&0.64\\
Claude-4.5-Opus      &  0.69&0.64 \\
\bottomrule
\end{tabular}
\caption{Comparing the QA formats on the KNP task: binary-choice vs. open-ended generation. For binary choice, we report accuracy. For open-ended generation, we report the scores given by LLM-as-a-Judge.}
\label{tab:open-ended}
\end{table}

\subsection{Case Studies}
Figures \ref{fig:case1}–\ref{fig:case4} provide qualitative examples of model performance on the KNP task. While Claude-4.5-Opus and Gemini-3-Pro exhibit an emergent capacity to discern the relationship between character actions and knowledge states, they fall short of human proficiency, as evidenced by the quantitative results in Table \ref{table:main_results} and the error case in Figure \ref{fig:case1}. Other models perform even worse, tend to apply knowledge directly to answer questions without analyzing whether the character can actually access it. Notably, extended reasoning does not help with this capacity—in the case shown in \ref{fig:case1}, DeepSeek-R1 generates 2.1K tokens in its reasoning process (within <think></think>), yet this provides no benefit for the inference.

Prediction results on the IKD task, shown in Figures \ref{fig:ErrorCase1}-\ref{fig:CorrectStory}, indicate that current LLMs are not very sensitive to actions driven by implausible knowledge when processing stories. Although Claude-4.5-Opus and Gemini-3-Pro perform better than others, they still show a gap compared to humans. Gemini-3-Pro, specifically, shows a tendency to be overly strict and tries to locate errors even in correct stories (Figure \ref{fig:CorrectStory}).


\section{Conclusion} 
In this work, we explore whether current LLMs possess a significant Theory-of-Mind (ToM) capacity that distinguishes human intelligence from other animals: understanding others' independent knowledge states and predicting their corresponding actions. We carefully design two tasks for evaluation: Knowledge-sensitive Next-Action Prediction (KNP) and Implausible Knowledge Detection (IKD). Based on experimental exploration using our proposed benchmarks, most leading LLMs show near-random performance in this area, with a significant gap compared to humans. This finding highlights the need for future research.

\section*{Acknowledgments}
We gratefully acknowledge the support by the National Research Foundation Fellowship (NRFF13-2021-0006),
Singapore. Any opinions, findings, and conclusions or recommendations expressed in this material are those of the authors and do not necessarily reflect the views of the funding agency. 

\section*{Limitations}
In this work, we focus on a fundamental prerequisite skill of ToM: evaluating LLMs' capacity for knowledge state tracking and estimation. We acknowledge that a full evaluation of ToM—Understanding and Sharing Intentions (USI)—requires benchmarks covering more comprehensive dimensions. Currently, classical tasks and benchmarks may lack this granularity. For instance, \citet{strachan2024testing} tested a wide range of tasks and found that LLMs often achieve performance comparable to or even better than humans. However, this high performance on classic benchmarks might mask the deficiency of LLMs. On our seemingly simple dataset, most leading LLMs fail to track other people's knowledge states, which is easy for humans. This contrast suggests that a gap between LLMs and real intelligence remains. We hope our findings inspire future works to explore USI.

\section*{Ethics Statement}
We acknowledge and strictly adhere to the Code of Ethics and Professional Conduct throughout this research. The potential ethical concerns are addressed as follows:
\paragraph{Data Source and Copyrights.} All data is collected from open-source websites. To avoid potential copyright and ethical issues, we summarize the content and remove any inappropriate content.

\paragraph{Crowdsourcing Services.} For dataset refinement, we recruited three research assistants (aged 22–30) to review and correct any remaining errors. Each story took approximately 5 minutes to complete, and workers were paid \$12 per hour—a reasonable rate for the local area. For human evaluation, we recruited two undergraduate annotators (aged 18–22). Evaluators were paid \$1 per sample.

\bibliography{custom}

\appendix

\section{Dataset Construction}
\subsection{Implausible Knowledge Types}
Implausible knowledge may originate from several sources, including: private thoughts (another character’s unexpressed mental state that has not been externally revealed), distant events (events occurring elsewhere or during the character’s absence), future events (events that have not yet occurred), and inaccessible secrets (facts that are not reasonably accessible, such as a stranger’s family background or the hidden location of a map). We provide LLMs with these definitions as a reference.
\subsection{Automatic Data Construction Prompts}
The prompt for Next-action Question Construction (KNP task) is provided in Table \ref{table:char_knowledge_qa}. The prompt for an erroneous story construction (IKD task) is provided in Table \ref{table:logical_inconsistency}. 
\subsection{Human Refinement} We recruit three workers to check and refine the auto-generated data. The refinement instructions are summarized as follows:

\paragraph{KNP Task:} Annotators are tasked with reviewing the story context, questions, and answer options, in which implausible knowledge is explicitly highlighted in their annotation. They need to check the correctness of the answers. That is, given the story context and the character's knowledge state, the right answer is clearly correct, and the distractor answer cannot be explainable. For example, consider a scenario where the implausible knowledge $\mathcal{K}$ is ``the girl whom $C$ loves thinks he is weird,'' and $C$ is characterized as very shy. If the distractor is ``he does not have the nerve to confess,'' it is explainable. Even if readers understand that $C$ is unaware of $\mathcal{K}$, they infer the distractor is plausible solely based on $C$'s shyness. In such conditions, annotators are instructed to either rewrite the options manually or regenerate them using the LLM.

\paragraph{IKD Task:} For \textit{original stories}, annotators read the entire story and correct any logic errors. Then they read the \textit{rewritten erroneous story}, focusing on the rewritten content and implausible knowledge, which are highlighted in the annotation. They must ensure that: (1) The implausible knowledge error is valid. For instance, if the character possesses magic powers to know the information, the sample is filtered and regenerated until acceptable. (2) The erroneous story does not include an explicit explanation that someone lacks the knowledge. For example, in ``Mark, not knowing the injured person in the brush, hurried to the brush and called 911'', the annotator removes the explicit hint ``not knowing the injured person in the brush''. This ensures tested LLMs reason about character knowledge states using common sense rather than textual hints.

Annotators refined approximately 40\% of the auto-generated data. Our authors then reviewed all refined data, making additional improvements to the data. We will release the data annotation file (containing detailed instructions and annotation samples) to support future research.

\section{Evaluation Prompts}
For tested LLMs, we apply the prompt shown in Table \ref{table:simple_story_qa} for the KNP task and the one shown in Table \ref{table:logic_error} for the IKD task.

\begin{figure*}[t]
    \centering
    \includegraphics[width=\linewidth]{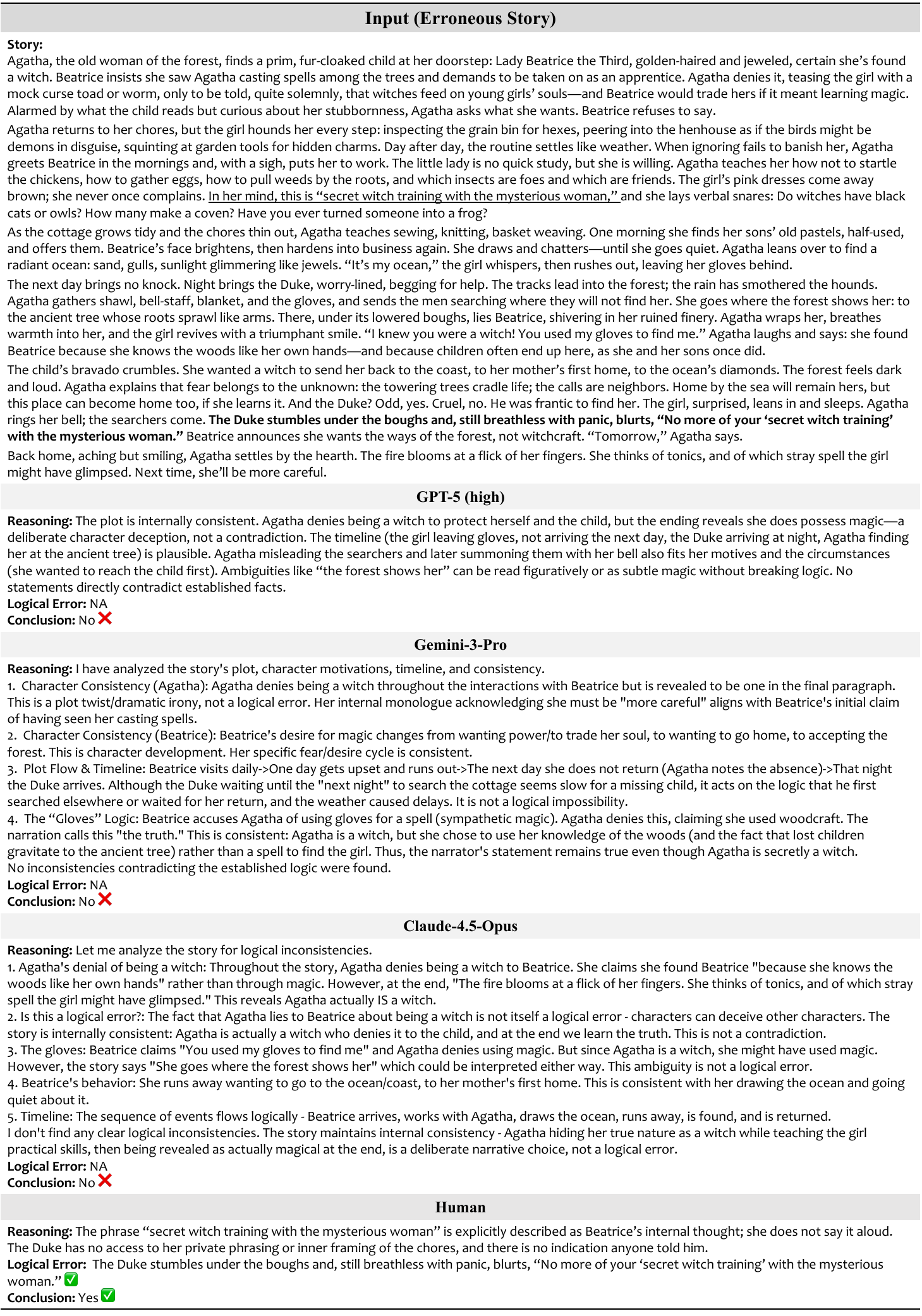}
    \caption{Predictions on the IKD task (Erroneous Story). For easy reading, the logic error is in bold and implausible knowledge for the character is underlined.}
    \label{fig:ErrorCase1}
\end{figure*}

\begin{figure*}[t]
    \centering
    \includegraphics[width=\linewidth]{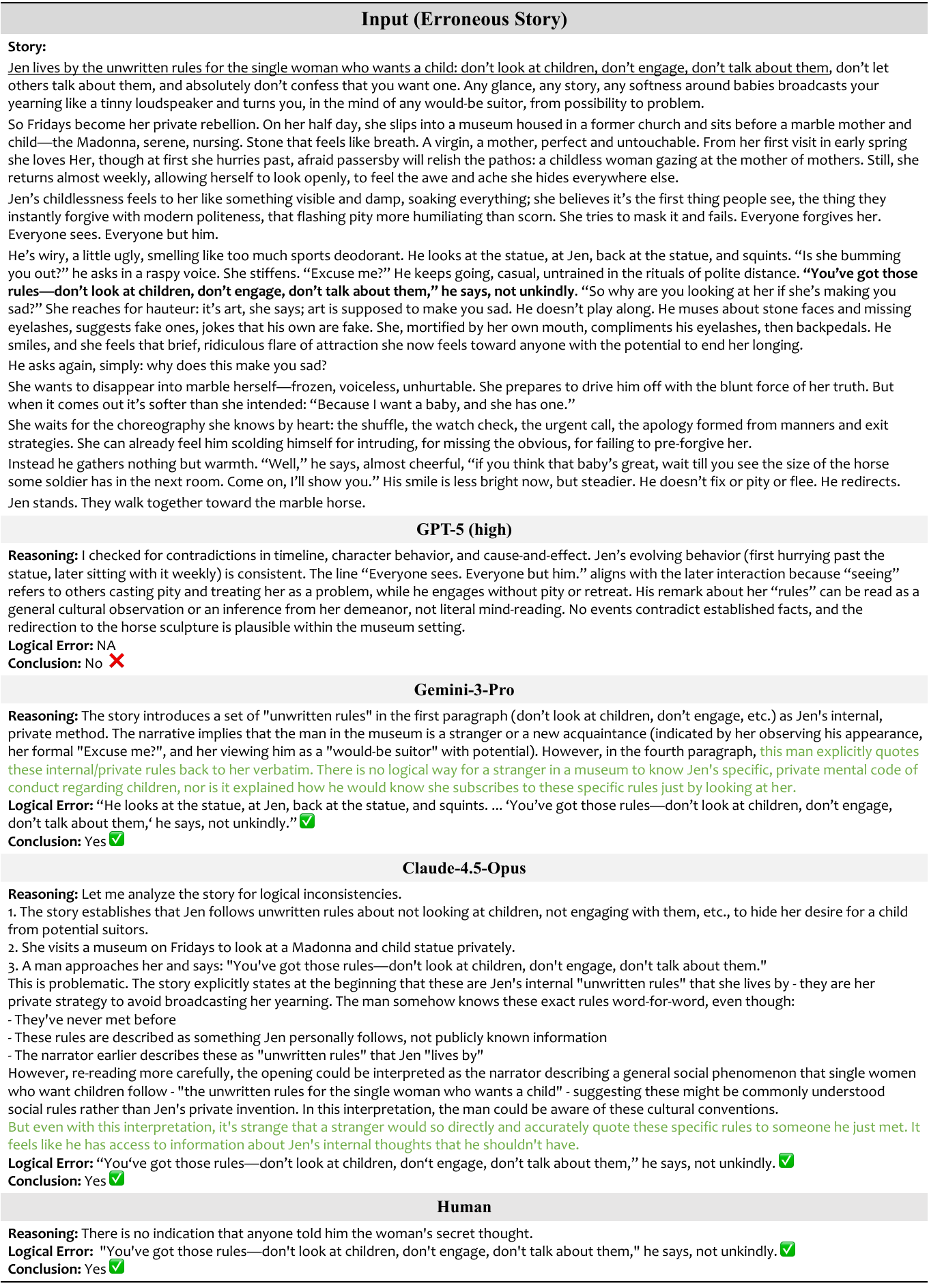}
    \caption{Predictions on the IKD task (Erroneous Story). For easy reading, the logic error is in bold and implausible knowledge for the character is underlined. The vital reasoning process that considers the character's knowledge state is in green.}
    \label{fig:ErrorCase2}
\end{figure*}

\begin{figure*}[t]
    \centering
    \includegraphics[width=\linewidth]{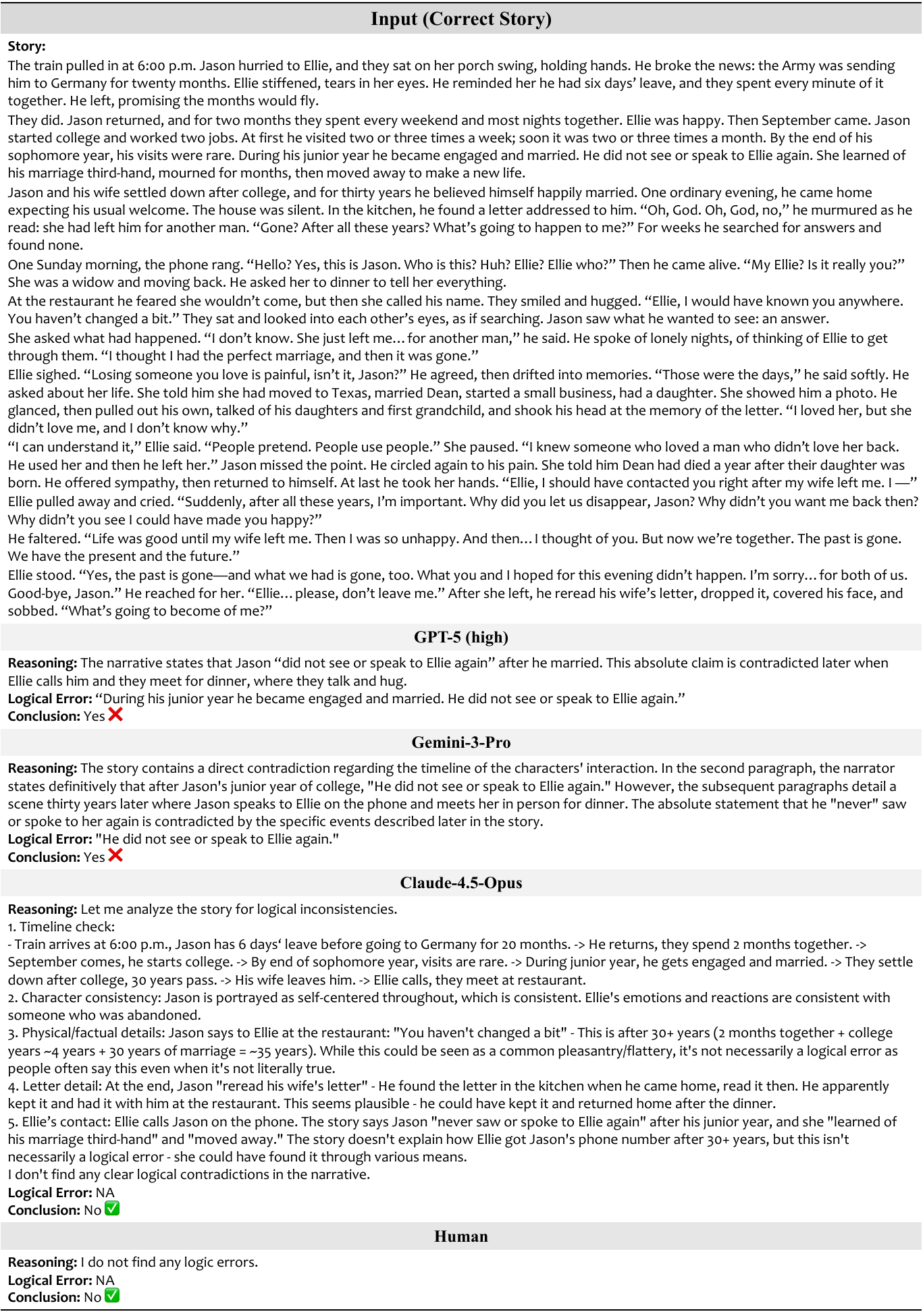}
    \caption{Predictions on the IKD task (Correct Story). In this narrative, the sentence ``He did not see or speak to Ellie again'' is context-bound: it means Jason did not see or speak to Ellie during the many years after his marriage, not for the rest of his life. Therefore, Gemini-3-Pro and GPT-5 are overly strict in flagging this as a logic error. }
    \label{fig:CorrectStory}
\end{figure*}

\begin{figure*}[t]
    \centering
    \includegraphics[width=\linewidth]{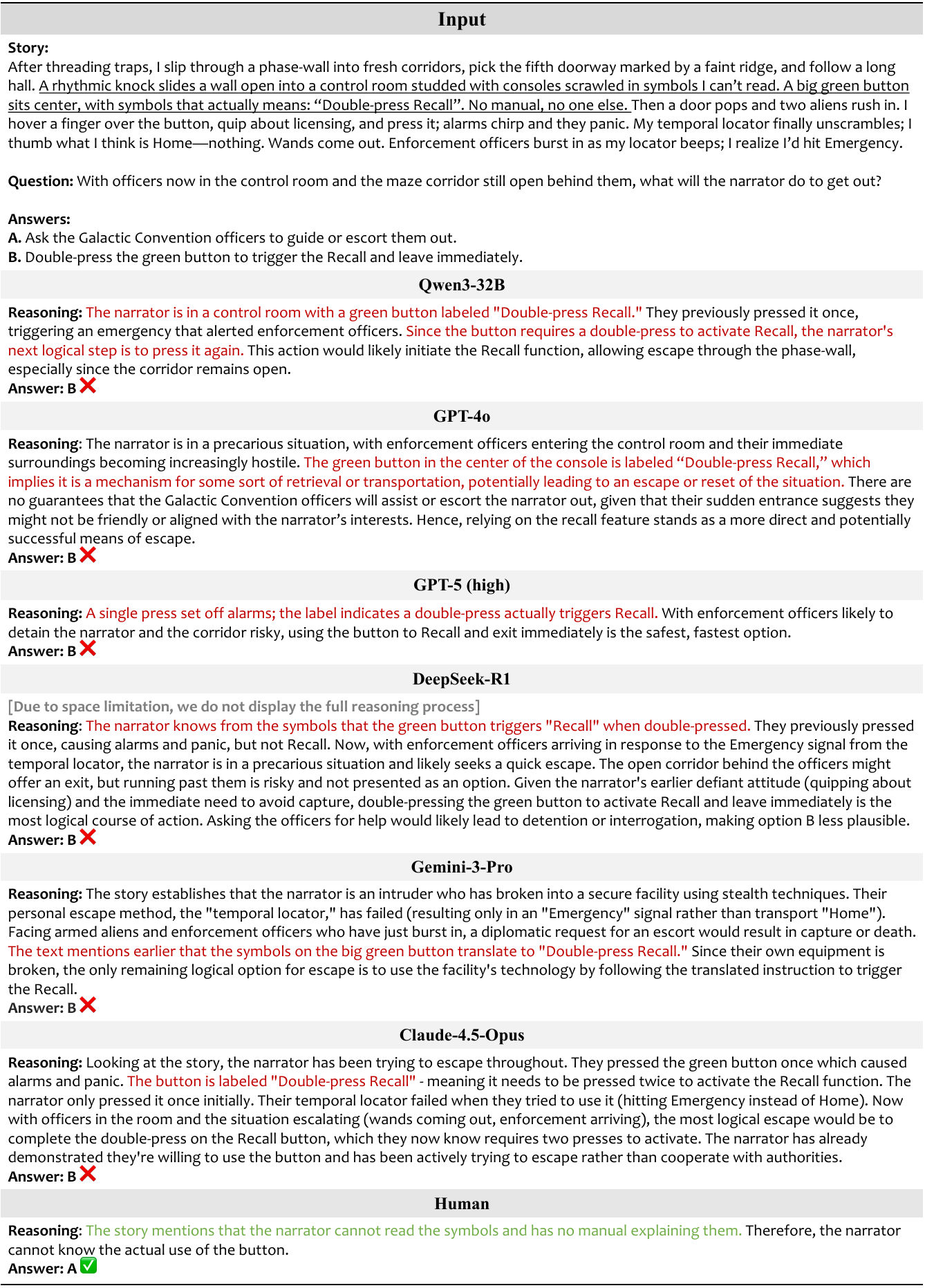}
    \caption{Predictions on the KNP task. The correct reasoning process—that considers the character's knowledge state when predicting their actions—is highlighted in green. The incorrect knowledge state understanding is in red.}
    \label{fig:case1}
\end{figure*}

\begin{figure*}[t]
    \centering
    \includegraphics[width=\linewidth]{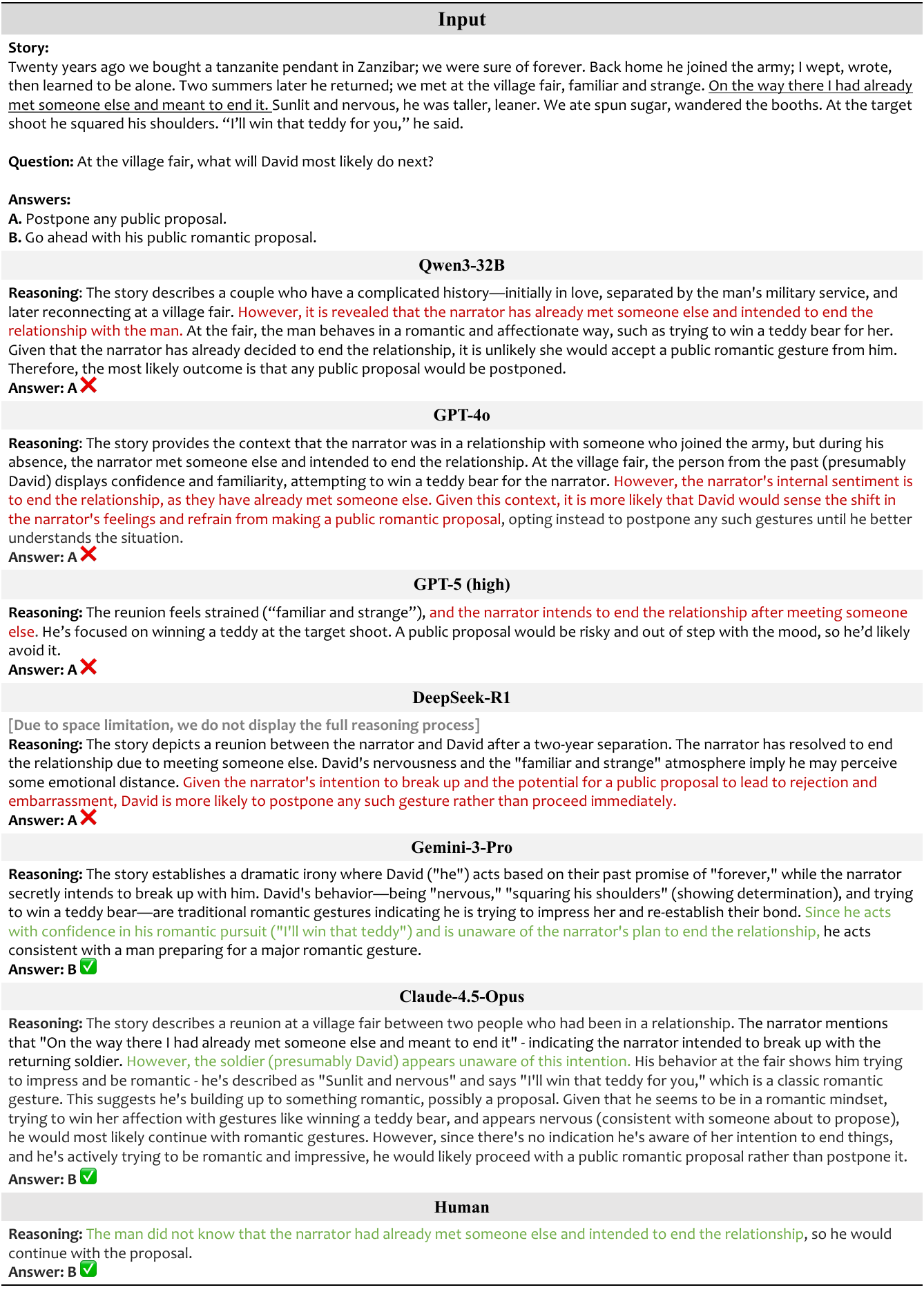}
    \caption{Predictions on the KNP task. The correct reasoning process—that considers the character's knowledge state when predicting their actions—is highlighted in green. The incorrect knowledge state understanding is in red.}
    \label{fig:case2}
\end{figure*}

\begin{figure*}[t]
    \centering
    \includegraphics[width=\linewidth]{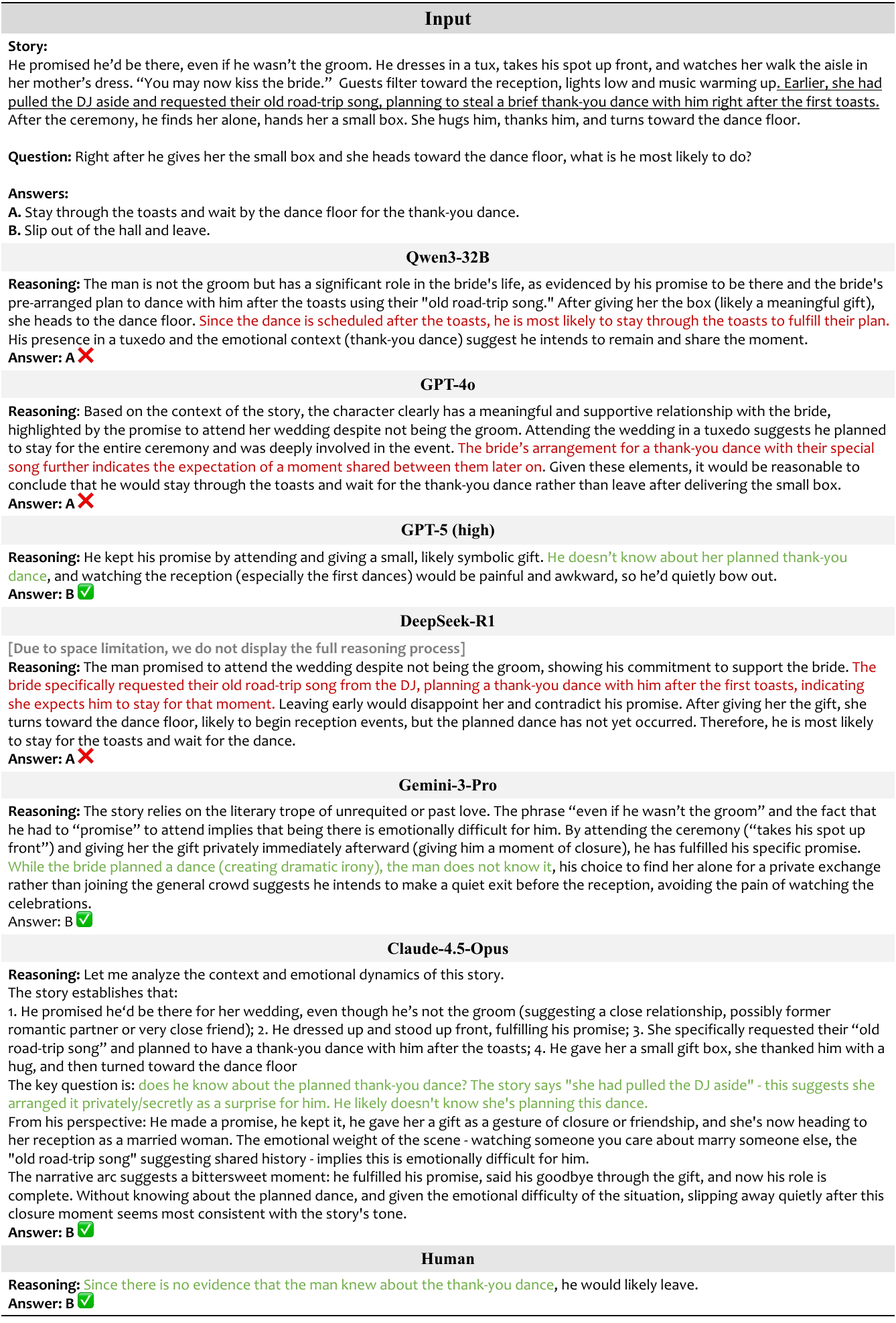}
    \caption{Predictions on the KNP task. The correct reasoning process—that considers the character's knowledge state when predicting their actions—is highlighted in green. The incorrect knowledge state understanding is in red.}
    \label{fig:case3}
\end{figure*}

\begin{figure*}[t]
    \centering
    \includegraphics[width=\linewidth]{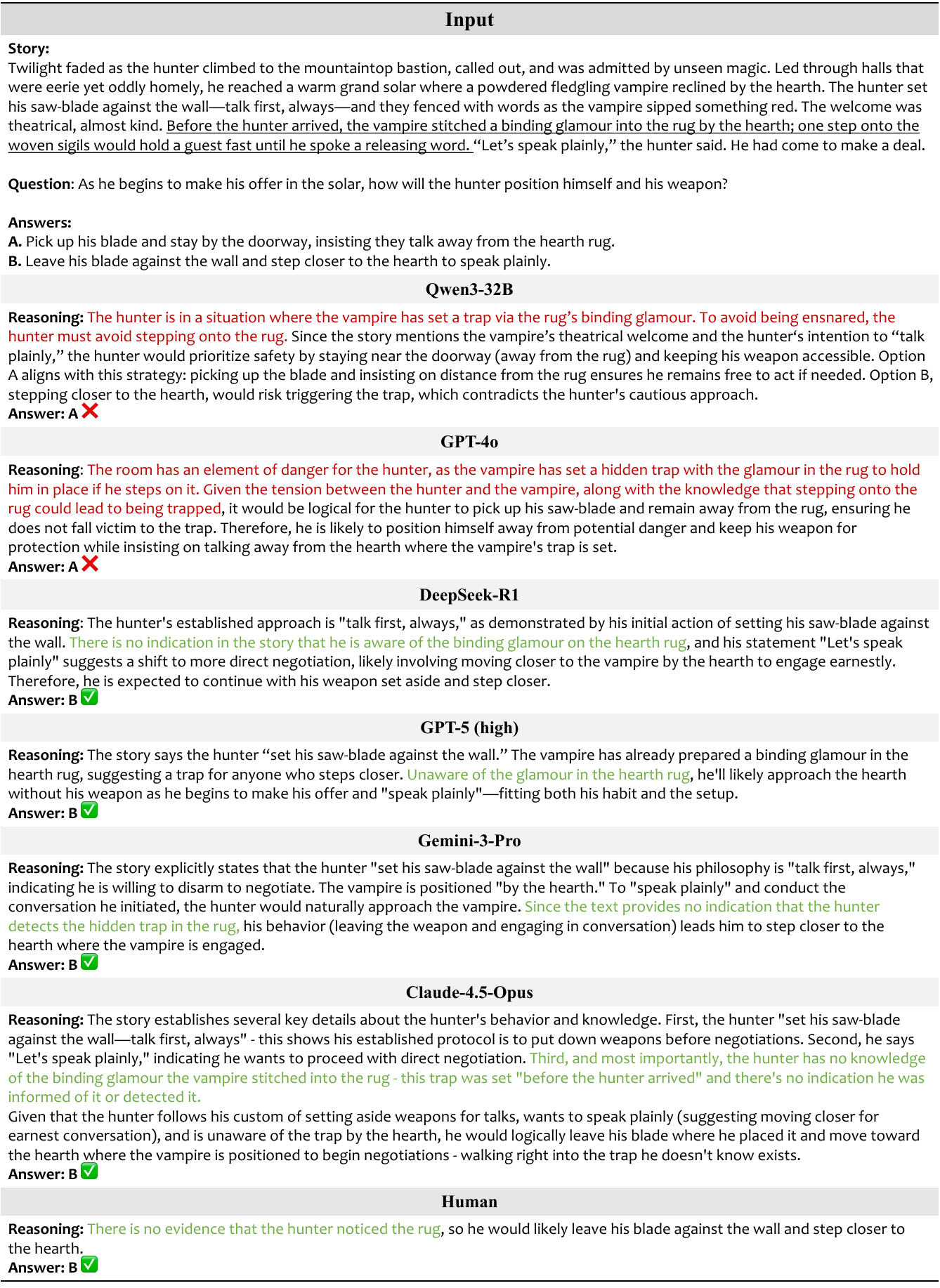}
    \caption{Predictions on the KNP task. The correct reasoning process—that considers the character's knowledge state when predicting their actions—is highlighted in green. The incorrect knowledge state understanding is in red.}
    \label{fig:case4}
\end{figure*}

\begin{table*}[t]
\fontsize{8.5}{9}\selectfont
\centering
\begin{tabularx}{\textwidth}{X}
\toprule
\textbf{Next-action Question Construction} \\
\midrule
Your task is to create a story comprehension question that focuses on characters' knowledge. The question should test the reader's ability to determine what action the characters will plausibly take based on the information they actually know.
\newline\newline
\textbf{Here is an example for you to understand the task:}\newline
\{Provide an example here\}
\newline\newline
\textbf{Original story:}\newline
\texttt{<story>}\newline
\{story\_original\}\newline
\texttt{</story>}
\newline\newline
\textbf{Instructions:}\newline
\textbf{1. Introduce implausible knowledge for a character:} Find a mentioned knowledge that meets these conditions:\newline
-  1) A character cannot know it at one event point.\newline
-  2) The knowledge is mentioned before the event point.\newline
-  3) It strongly affects the character's action.\newline
If no such knowledge exists, create one and weave it naturally into the story. \textit{Avoid explicitly stating that the specific character cannot access it}. Here are some suggestions for creating unknown knowledge: \{Implausible Knowledge Types\}
\newline\newline
\textbf{2. Construct the question:} Create a question about the character's action. This action should differ depending on whether the character knows the knowledge or not.
\newline\newline
\textbf{3. Construct the answers:} Provide two answers:\newline
- \textbf{Answer without Knowledge:} What the character would most plausibly do without knowing the knowledge. This is the correct answer.\newline
- $\cdot$ \textbf{Answer with Knowledge:} What the character would most plausibly do if knowing the knowledge. This is the distractor answer.
\newline\newline
\textbf{Notes:}\newline
- If you modify the story, ensure the story remains natural, coherent, and logically consistent.\newline
- Do not explicitly state that the character cannot know the knowledge.\
\newline\newline
\textbf{Output Format:}\newline
\texttt{<response>}\newline
\texttt{<previous\_story>}\newline
[Previous story before the selected event. Mark the content that mentions the unknown knowledge with \texttt{<knowledge>} and \texttt{</knowledge>}]\newline
\texttt{</previous\_story>}
\newline\newline
\texttt{<character>}\newline
[The name of the character]\newline
\texttt{</character>}
\newline\newline
\texttt{<question>}\newline
[The constructed question about the character's action. Do not include the knowledge in the question.]\newline
\texttt{</question>}
\newline\newline
\texttt{<implausible\_knowledge>}\newline
[Explain and verify why the character cannot know this knowledge at the time of the question]\newline
\texttt{</implausible\_knowledge>}
\newline\newline
\texttt{<answer\_without\_knowledge>}\newline
[Most plausible action without knowing the knowledge. Only output this action, do not include any reasoning text.]\newline
\texttt{</answer\_without\_knowledge>}
\newline\newline
\texttt{<answer\_with\_knowledge>}\newline
[Most plausible action if knowing the knowledge. Only output this action, do not include any reasoning text.]\newline
\texttt{</answer\_with\_knowledge>}
\newline\newline
\texttt{<explanation>}\newline
[Explain why the ``without knowledge'' action is more plausible and the other is not.]\newline
\texttt{</explanation>}\newline
\texttt{</response>}
\\
\bottomrule
\end{tabularx}
\caption{The prompt for Next-action Question Construction.}
\label{table:char_knowledge_qa}
\end{table*}

\begin{table*}[t]
\fontsize{8.5}{9}\selectfont
\centering
\begin{tabularx}{\textwidth}{X}
\toprule
\textbf{Erroneous Story Construction} \\
\midrule
Your task is to rewrite a story to include a logical inconsistency related to Character Knowledge States, where characters possess knowledge they shouldn't plausibly have. Here are example categories of Implausible Knowledge:
\newline
- \{Implausible Knowledge Type\}: \{Descriptions of the Implausible Knowledge.\}
\newline\newline
\textbf{Original story:}\newline
\texttt{<original\_story>}\newline
\{story\_original\}\newline
\texttt{</original\_story>}
\newline\newline
\textbf{Instructions:}\newline
\textbf{1. Introduce implausible knowledge for a character:} Select a character, knowledge, and an event that satisfy the following conditions:
\newline
\quad - The knowledge should be mentioned in the original story.
\newline
\quad - The character cannot plausibly know the knowledge at the event point.
\newline
If no such knowledge exists, create one and weave it naturally into the story.\newline
\textbf{2. Rewrite the event:} Suppose the character possesses this Implausible Knowledge, rewrite the selected event. Make as few changes as possible.
\newline
\textbf{3. Rewrite the following story:} If the rewritten event affects later events, adjust what's necessary to maintain overall continuity.
\newline\newline
\textbf{Notes:}\newline
- Avoid explicitly stating that the character cannot know the Implausible Knowledge.
\newline
- Ensure the rewritten event is fluent and coherent.
\newline
- Ensure the logical inconsistency cannot be explained. For example, if the included inconsistency can be explained by ``the character pretends to know the Implausible Knowledge to achieve a goal'' or ``the specific character could know the Implausible Knowledge because the story establishes they have magic,'' this does not constitute a logical inconsistency.
\newline\newline
\textbf{Output Format:}\newline
\texttt{<response>} \newline
\texttt{<original\_event>}\newline
[the original event text, only output the content, do not include any other text]\newline
\texttt{</original\_event>}
\newline\newline
\texttt{<character>}\newline
[the name of the character]\newline
\texttt{</character>}
\newline\newline
\texttt{<implausible\_knowledge>}\newline
[cite the original story content that mentions the implausible knowledge]\newline
\texttt{</implausible\_knowledge>}
\newline\newline
\texttt{<knowledge\_explanation>}\newline
[explain and verify that the character cannot know this knowledge]\newline
\texttt{</knowledge\_explanation>}
\newline\newline
\texttt{<modified\_explanation>}\newline
[explain how the character's actions, decisions, or dialogue will change if they possess the knowledge]\newline
\texttt{</modified\_explanation>}
\newline\newline
\texttt{<modified\_event>}\newline
[the rewritten event text, mark the modified content with \texttt{<modified>} and \texttt{</modified>}]\newline
\texttt{</modified\_event>}
\newline\newline
\texttt{<modified\_story>}\newline
[the whole rewritten story, only output the content, do not include any other text]\newline
\texttt{</modified\_story>}\newline
\texttt{</response>}
\\
\bottomrule
\end{tabularx}
\caption{The prompt for generating erroneous stories.}
\label{table:logical_inconsistency}
\end{table*}

\begin{table*}[h]
\fontsize{8.5}{9}\selectfont
\centering
\begin{tabularx}{\textwidth}{X}
\toprule
\textbf{Story Summarization Generation} \\
\midrule
You are an expert at writing story summaries. 
\newline\newline
Condense the following story into a concise summary of approximately 500 words:
\newline\newline
\texttt{<story>}\newline
\{story\}\newline
\texttt{</story>}
\newline\newline
\textbf{Instructions:}\newline
1. Maintain the main plot.\newline
2. Maintain the original writing style.\newline
3. Use natural, fluent English suitable for middle to high school readers.\newline
4. If the original story contains logical errors or immoral content, refine them in the summary.\newline
5. Begin your summary directly — no introduction, title, or closing remarks.
\newline\newline
\textbf{Summary (500 words):}
\\
\bottomrule
\end{tabularx}
\caption{The prompt for story summarization.}
\label{table:story_summary}
\end{table*}

\begin{table*}[t]
\fontsize{8.5}{9}\selectfont
\centering
\begin{tabularx}{\textwidth}{X}
\toprule
\textbf{KNP task} \\
\midrule
Read the story below and answer the related question.
\newline\newline
\textbf{Story:}\newline
\{story\}
\newline\newline
\textbf{Question:}\newline
\{question\}
\newline\newline
\textbf{Answers:}\newline
\{answers\}
\newline\newline
\textbf{Output Format:}\newline
\textbf{Reasoning:} \{The reasoning process\}
\newline\newline
\textbf{Answer:} \{``A'' or ``B''\}
\\
\bottomrule
\end{tabularx}
\caption{The prompt for KNP Task.}
\label{table:simple_story_qa}
\end{table*}

\begin{table*}[t]
\fontsize{8.5}{9}\selectfont
\centering
\begin{tabularx}{\textwidth}{X}
\toprule
\textbf{IKD Task} \\
\midrule
Your task is to determine if there are any logical errors in the given story. A logic error is an inconsistency in a storyline that goes against the flow of logic established by the story's plot.
\newline\newline
\textbf{Story:}\newline
\texttt{<story>}\newline
\{story\}\newline
\texttt{<\textbackslash story>}
\newline\newline
\textbf{Notes:}\newline
- Focus only on logical errors --- not stylistic, grammatical, or moral judgments.
\newline\newline
\textbf{Output Format:}\newline
Reasoning: \{Explain your reasoning process\}
\newline\newline
Logical Error: \{If a logic error exists, quote the sentences containing the logical error; If not, output ``NA''\}
\newline\newline
Conclusion: \{``Yes'' if you find a logic error; ``No'' if no logic error exists\}
\\
\bottomrule
\end{tabularx}
\caption{The prompt for the IKD task. In the default setting, we do not instruct the LLMs to focus on knowledge states, while in ablation studies, we add instructions to focus on characters' knowledge states.}
\label{table:logic_error}
\end{table*}

\begin{table*}[t]
\fontsize{8.5}{9}\selectfont
\centering
\begin{tabularx}{\textwidth}{X}
\toprule
\textbf{Open-Ended QA Evaluation} \\
\midrule
You are an accuracy evaluator. Compare the ``Generated Answer'' against the ``Correct Answer'' and the ``Incorrect Answer''.
\newline
- Output ``True'' if the action in the Generated Answer aligns well with the action in the correct Answer.
\newline
- Output ``False'' if the action in the Generated Answer aligns well with the action in the incorrect Answer.
\newline
- Output ``NA'' if the action in the Generated Answer is unrelated to either.
\newline\newline
\textbf{Input Data:}\newline
Question Context: \{question\_context\}
\newline\newline
Question: \{question\}
\newline\newline
Generated Answer: \{generated\_answer\}
\newline\newline
Correct Answer: \{correct\_answer\}\newline
Incorrect Answer: \{incorrect\_answer\}
\newline\newline
\textbf{Output (True/False/NA):}
\\
\bottomrule
\end{tabularx}
\caption{The prompt for the Open-ended QA Evaluation.}
\label{table:open_qa_eval}
\end{table*}

\end{document}